\pdfoutput=1

\documentclass[10pt,twocolumn,letterpaper]{article}

\usepackage[pagenumbers]{wacv} 

\usepackage[accsupp]{axessibility}
\usepackage{graphicx}
\usepackage{amsmath}
\usepackage{amssymb}
\usepackage{booktabs}
\usepackage{microtype}
\usepackage{float}
\usepackage{multirow}
\usepackage{import}
\usepackage{xifthen}
\usepackage{tikz}
\usepackage{svg}
\usepackage{array}
\newcolumntype{P}[1]{>{\centering\arraybackslash}p{#1}}
\usepackage{makecell}

\usepackage[pagebackref,breaklinks,colorlinks]{hyperref}

\usepackage[capitalize]{cleveref}
\crefname{section}{Sec.}{Secs.}
\Crefname{section}{Section}{Sections}
\Crefname{table}{Table}{Tables}
\crefname{table}{Tab.}{Tabs.}

\def\wacvPaperID{517} 
\def\confName{WACV}
\def\confYear{2024}

\begin{document}

\title{Towards Addressing the Misalignment of Object Proposal Evaluation for Vision-Language Tasks via Semantic Grounding}

\author{Joshua Feinglass and Yezhou Yang\\
Active Perception Group, Arizona State University\\
{\tt\small \{joshua.feinglass,yz.yang\}@asu.edu}
}
\maketitle
\begin{abstract}
Object proposal generation serves as a standard pre-processing step in Vision-Language (VL) tasks (image captioning, visual question answering, etc.). The performance of object proposals generated for VL tasks is currently evaluated across all available annotations, a protocol that we show is ``misaligned'' - {\it  higher scores do not necessarily correspond to improved performance on downstream VL tasks}. Our work serves as a study of this phenomenon and explores the effectiveness of semantic grounding to mitigate its effects. To this end, we propose evaluating object proposals against only a subset of available annotations, selected by thresholding an annotation importance score. Importance of object annotations to VL tasks is quantified by extracting relevant semantic information from text describing the image. We show that our method is consistent and demonstrates greatly improved alignment with annotations selected by image captioning metrics and human annotation when compared against existing techniques. Lastly, we compare current detectors used in the Scene Graph Generation (SGG) benchmark as a use case, which serves as an example of when traditional object proposal evaluation techniques are misaligned\footnote{Source codes, data, and surveys will be released at \url{https://github.com/JoshuaFeinglass/VL-detector-eval}.}.
\end{abstract}
\section{Introduction}
\begin{figure}[t!]
\centering
\def\svgwidth{\columnwidth}
\includegraphics[width=\columnwidth,height=\textheight,keepaspectratio]{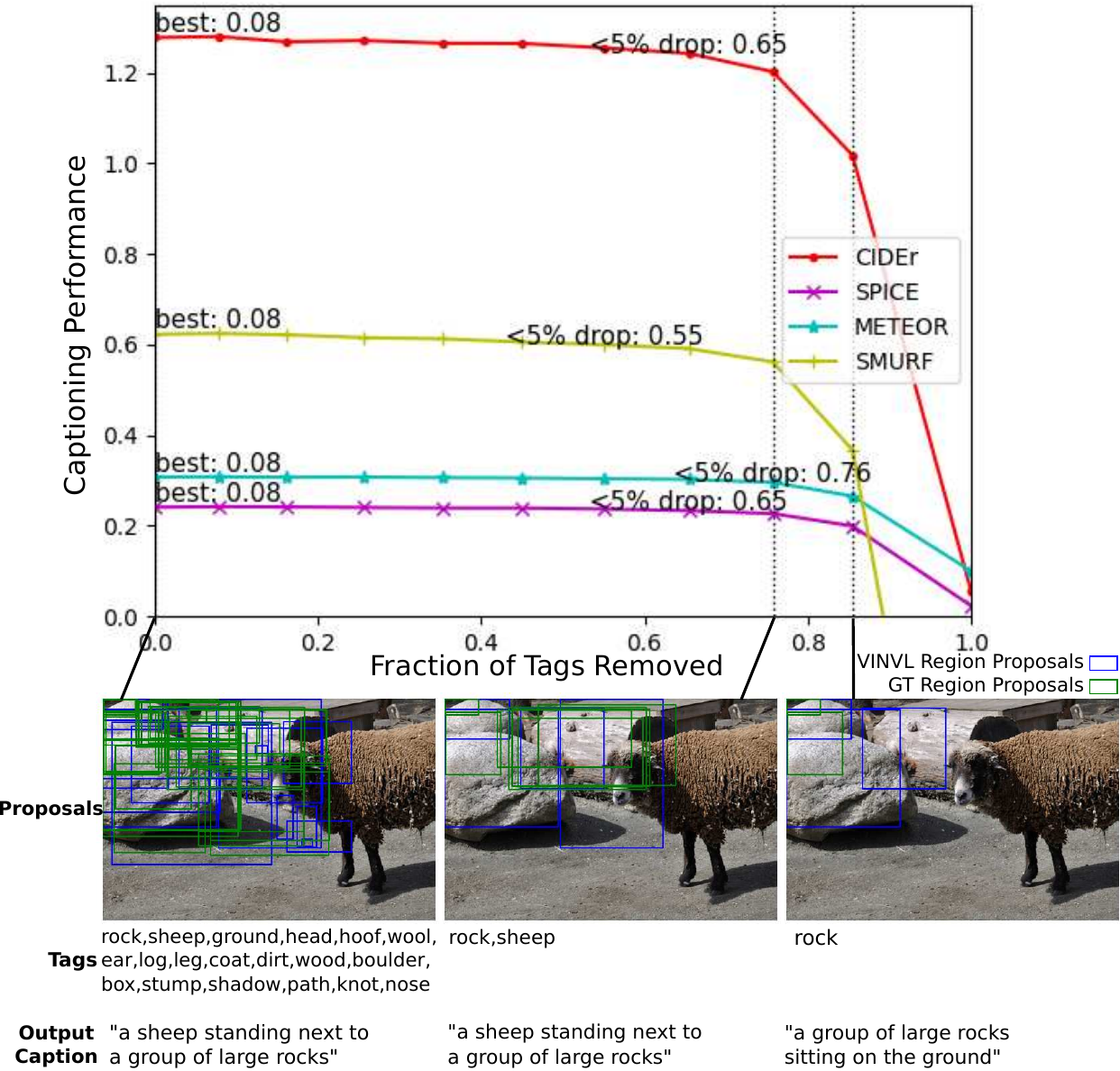}
\caption{A system-level plot showing the performance of OSCAR \cite{oscar2020} as tags and their corresponding image features are removed based on our proposed annotation importance scores. Captions are evaluated using standard metrics \cite{cider2015,meteor2005,spice2016,smurf2021} with all punctuation removed. The text on the plot shows at what fraction of tag and corresponding bounding box feature removal the metrics achieve their best score and the highest fraction of tags that can be removed before the performance drops by more than 5\% for each metric. The results suggest that model performance depends on a critical subset of object regions.}
\label{main_figure}
\end{figure}
\par
Vision-Language (VL) tasks are a growing topic in both the Natural Language Processing (NLP) and Computer Vision communities with the majority of techniques relying on object proposal generation for pre-processing \cite{topbottom2018,vinvl2021}. Object proposals are a set of regions or bounding boxes deemed likely to contain the object specified by a detector. Object proposal generation offers an explainable, efficient, and highly effective bridge between raw images and VL tasks. 
\par
However, current evaluation techniques of object proposal generation are poorly aligned with the VL use-case, resulting in adverse effects like ``gameability'' \cite{gameable2016}. While \cite{gameable2016} claim that this misalignment is caused by missing annotations, we theorize that the inclusion of superfluous object annotations not relevant to VL tasks in evaluation is also a contributing factor. Contrary to the prevailing attitude that models should be evaluated across all available annotations, we postulate that models only need information about a few critical objects to understand a scene. This intuition aligns well with the idea that not all test examples are equally important for evaluation \cite{notequal2021}, which is rapidly gaining traction in NLP benchmarks and benefits not just evaluation but data annotation as well. Thus, we propose selecting ground truth annotations for use in evaluation based on a semantic grounding signal, specifically image captions or region descriptions. To measure the importance of a given object, we extract relevant semantic information using typicality analysis \cite{smurf2021,typicality1997} and propagate this importance to adjacent objects using graph signal processing techniques \cite{gsp2017}. This importance score is then used to select only the objects most relevant to VL tasks for evaluation. Exploring image captioning as a case study of this phenomenon, we observe that {\it high image captioning performance can be maintained with only 24\%-44\% of object tags and their corresponding features from regions of interest depending on the performance metric} as shown in Figure \ref{main_figure}. Furthermore, the preservation of a high SMURF \cite{smurf2021} score suggests that removing these annotations does not significantly impact the detail/diversity of the generated captions.  
\par
Due to the scarcity of relevant detectors and lack of related benchmarks, we opt for a holistic approach when validating our metric. We perform three independent studies, each of which provides unique insight into the effectiveness and advantages of our approach. We begin with an empirical analysis and find that annotations selected using our importance scoring result in the highest alignment with improved image captioning performance for a widely adopted captioning pipeline when compared to an area-based baseline. To get an example-level view, we perform three human surveys using Amazon Mechanical Turk (AMT) and find that our proposed metric adjustments are highly aligned with human judgement while most of the existing metrics exhibit little to no alignment with VL bias. We then show that our selections are consistent across human-annotated text descriptions from different datasets, in particular, COCO captions and Visual Genome region descriptions. Our findings support the existence of a critical annotation set which remains consistent when using different semantic grounding sources. Furthermore, in our last experiment, we explore a Scene Graph Generation (SGG) use case where our approach provides information about model performance not captured in previous benchmarks. More specifically, we observe an instance where the standard evaluation approach fails to capture poor precision performance on VL task essential objects due to its misalignment. \vspace{1mm}
\\
\textbf{Contributions:} We create a theoretical formulation of misalignment in object proposal evaluation and develop an object importance score which can be used to mitigate the effect of this phenomenon and enhance the feedback provided when designing VL detectors. To support these insights, we perform 4 experiments: an analysis of the alignment between our importance score and performance on a downstream VL task, 3 human surveys, a study of the consistency between selected object regions from different annotation sets, and a demonstration of mitigated misalignment on a SGG benchmark.
\section{Related Work}
\noindent
\textbf{Object Proposal Evaluation} relies primarily on variations of mean Average Precision (mAP) \cite{cocoeval2015,evalreview2020,pascal2015,generalizedeval2019}, although Average Recall (AR) and mAR (mean Average Recall) are employed for evaluation benchmarks of related tasks like SGG \cite{sgbase2017}. These methods are considered to be intuitive and are not validated against human judgement. There are no existing benchmarks or metrics for VL task related object detection, despite the existence of benchmarks for other sub-tasks like salient object detection \cite{salient2015}. Thus, our work is the first to introduce such a benchmark. There are however numerous scene-oriented object detectors developed via pre-training \cite{vinvl2021,graphical2019,motifs2018,unbiased2020,sgg2021,topbottom2018}, with Visual Genome (VG) \cite{vg2017} serving as the standard dataset. 
\\
\textbf{Scene Understanding} tasks including scene representation and scene recognition rely largely on supervisory signals such as object segmentation and labels, which can be erroneous or incomplete \cite{because2021}. Previous works have also shown that human captions and text alone can serve as a strong supervisory signal for object detection \cite{vocab2021,weak2021,sam2023} and Visual Question Answering (VQA) \cite{banerjee2021weaqa}. \cite{scene2021} sought to find the minimum set of objects needed for the task of scene recognition. Objects relations have also been shown to be important for scene representation and recognition \cite{relations2020}, with graph-based methods achieving significant success \cite{graphlearning2020}. Dataset filtering \cite{filter2019,filter2020} has also explored the use of supervisory signals for data example selection from an inference perspective. Our work combines these scattered concepts into a single coherent formulation of scene-oriented bias for evaluation.
\\
\textbf{Annotation Weighting} is gaining popularity with \cite{notequal2021} and \cite{right2019} asserting that each test example is not equally informative for evaluation benchmarks and that quantifying this importance can improve annotation and help detect overfitting. In particular, Item Response Theory (IRT) is a test example selection and weighting mechanism gaining popularity in Natural Language Processing benchmarks \cite{irt2016,irt2020,irt2021} which seeks to provide greater rewards for more difficult text examples and has been shown to be more reliable and representative than standard accuracy. Rather than selecting and weighting examples based on difficulty, our work instead focuses on selecting test examples based on their relevance to a task of interest.
\section{Our Approach}
\subsection{Vision-Language Task Background}
An object proposal based approach to VL tasks consists of a vision module \textbf{V} and cross-modal understanding module \textbf{VL}
\begin{equation}
\{v_d\}_{d \in \mathcal{D}}\!=\!\textbf{V}(image), \ \ y\!=\!\textbf{VL}(w_{task},\{v_d\}_{d \in \mathcal{D}}),
\end{equation}
where the pre-preprocessed image information $v_d=(b_d,f_d,c_d)$ consists of a region or bounding box $b_d$ and extracted features corresponding to the region $f_d$ along with a category label $c_d$ for each object detector proposal $d \! \in \! \mathcal{D}$. The text prompt $w_{task}$ and output $y$ are VL task-specific, corresponding to a question and answer in VQA, text and matching score in text-image retrieval, an empty prompt and an object predicate graph in SGG, and an empty prompt and output caption in image captioning. 
\subsection{Object Proposal Evaluation Background}
For a specific object category, an object region proposal $b_d$ provided by a detector is typically deemed correct or incorrect based on the largest intersection over union (IOU) it is able to achieve with a ground truth human region annotation $b_{a}$ from their category as shown
\begin{equation}
\text{IOU}(b_d,b_a) = \frac{area(b_d \cap b_{a})}{area(b_d \cup b_a)}.
\end{equation}
IOU is an extension of the Jaccard Index applied to a region's pixels. A threshold $\gamma$ is then applied to the IOU scores to obtain a set of correct detections. Precision is the most commonly used performance measure of an object detector proposal set $\mathcal{D}$ and is calculated over the ground truth region annotation set $\mathcal{A}$ for a specific category $\mathcal{A}_c \! \in \! \mathcal{A}$ as
\begin{equation}
P(\mathcal{D},\mathcal{A})=\frac{|\{max_{a \in \mathcal{A}_c}[\text{IOU}(b_d,b_a)] \geq \gamma\}_{d \in \mathcal{D}}|}{|\mathcal{D}|}.
\end{equation}
Further reading on object proposal evaluation is in \cite{evalsurvey2020}.
\subsection{Object Importance and Misalignment}
We now propose a novel formulation of object importance and explain how this results in metric misalignment. We first assume that the information relevant to task output $y$ provided by the output from the vision module $\textbf{V}$ is limited to a critical subset of ground truth object annotations $\mathcal{I} \! \in \! \mathcal{A}$. This implies that the increase in alignment, represented as the mutual information $MI$, between the output and $y$ provided by additional annotations is limited to an arbitrarily small constant $\delta$ such that
\begin{equation}
MI(\textbf{V}_{\mathcal{I}}(image);y)=MI(\textbf{V}_{\mathcal{A}}(image);y)+\delta.
\end{equation}
Thus, at a fixed number of detector proposals $|\mathcal{D}|$, a precision metric is misaligned when the ranking of detectors evaluated using the critical objects does not match the ranking of detectors evaluated using all the objects as shown
\begin{equation}
\label{eq:condition}
P(\mathcal{D}_1,\mathcal{I}) \! > \! P(\mathcal{D}_2,\mathcal{I}) \! \! \implies \! \! P(\mathcal{D}_1,\mathcal{A}) \! > \! P(\mathcal{D}_2,\mathcal{A}).
\end{equation}
Here $\mathcal{D}_1$ and $\mathcal{D}_2$ are object region proposals from two different detectors. More specifically, this condition is violated when ground truth region annotations from the outside the critical subset $\overline{\mathcal{I}}$ impact the ranking of the detectors by skewing the size of the correct detection set as shown
\begin{equation}
P(\mathcal{D}_1,\mathcal{I})+P(\mathcal{D}_1,\overline{\mathcal{I}}) < P(\mathcal{D}_2,\mathcal{I})+P(\mathcal{D}_2,\overline{\mathcal{I}}).
\end{equation}
This misalignment is more severe in tasks with a larger number of superfluous ground truth region annotations. By removing annotations that are unlikely to be critical to VL tasks, we reduce the size of $\overline{\mathcal{I}}$, thereby mitigating the risk of the condition from Eq.~\ref{eq:condition} being violated. Thus, we define an object's importance $I$ as the likelihood it is a member of $\mathcal{I}$.
\subsection{Estimating Object Importance}
\begin{figure}
\centering
\def\svgwidth{\columnwidth}
\includegraphics[width=\columnwidth,height=\textheight,keepaspectratio]{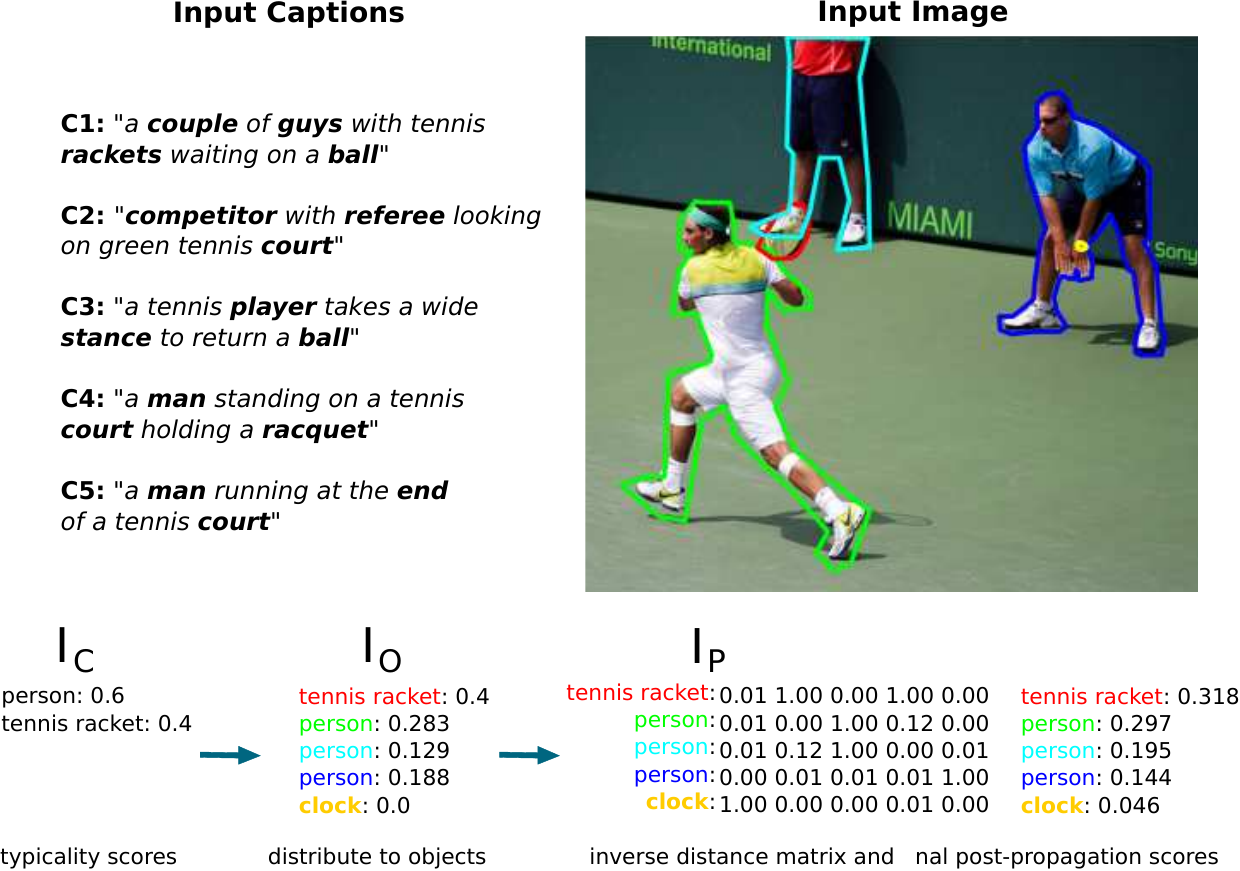}
\caption{An example illustrating our processing pipeline. Words used for object typicality are shown in bold font. Object annotations are color-coded to their corresponding label in the $I_O$ and $I_P$ processing stages. If we set $T=0.2$, only the tennis racket and adjacent player would be selected based on their high $I_P$ scores.}
\label{methodology-table}
\end{figure}
We estimate an object's importance using semantic grounding from human annotated captions for each image.  Our methodology consists of 3 steps: characterization of the underlying semantic process in order obtain importance scores for each object category present in the captions ($I_C$), distributing these importance scores to each object from the category based on the area of its region annotation ($I_O$), and propagating object importance to adjacent objects to reduce sparsity ($I_P$). Critical objects proposals to be used for evaluation are then selected based on a threshold $T$. An example is shown in Figure \ref{methodology-table}.\vspace{1mm}\\
\textbf{Typicality Scores ($I_C$)}: We utilize typicality \cite{smurf2021,typicality1997} to characterize the underlying semantic process generating the object instances present in the ground truth captions. To estimate the semantic typicality for our application, we extract the object-specific concepts from the caption using a Parts of Speech (POS) tagger \cite{spacy2017}. The prevalence of each object in the ground truth caption set $\mathcal{S}$ is then determined by taking its document frequency $df$ where each caption is treated as a separate document. The typicality is 
\begin{equation}
I_C(c_s)=\frac{df_{\mathcal{S}}(c_s)}{|\mathcal{S}|},
\end{equation}
where $c_s$ is an object category present in the caption sentence based on mappings from object-specific concepts to the most similar object class by ConceptNet \cite{conceptnet2018}, $|\mathcal{S}|$ is the number of captions present in the ground truth human caption annotations, $df_{\mathcal{S}}$ is a function that counts the number of captions in which the object-specific concept is present, and $I_C(c_s)$ is the estimated importance of the object category. In the rare case that no importance is assigned to any object categories, the data example has poor alignment between its captions and object annotations and is skipped during the evaluation.\vspace{1mm}\\
\textbf{Distribute to Objects ($I_O$)}:  To begin quantifying the importance of each object in the category $c_s$, the importance is then distributed to each of the ground truth object annotations from the given category $\mathcal{A}_{c_s}$ based on its area $area(b_a)$
\begin{equation}
I_O(b_a)=\frac{area(b_a)*I_{C}(c_s)}{\sum_{a \in \mathcal{A}_{c_s}}area(b_a)},
\end{equation}
where $I_O(b_a)$ is the importance of the object $b_a$.\vspace{1mm}\\
\textbf{Propogate Scores ($I_P$)}: We have now identified objects of importance to VL tasks in the image. However, larger scores are likely to be sparse among the objects since most objects are not in a category with a high $I_C$ score. We instead infer that VL task importance is highly dependent on regional interactions (e.g. person holding tennis rack in Figure \ref{methodology-table}) and utilize heat kernel based dispersion modeling techniques \cite{heat2008} from graph signal processing in order to capture inter-objects interactions. To this end, we model the objects in the image as a graph with adjacency matrix, $W$, and construct a heat kernel using the PyGSP toolkit \cite{gsp2017}. The values of $W$ are based on the inverse of the minimum Euclidean distance ($d$) between the ground truth object region annotations in the image as shown
\begin{equation}
W_{ij}=\frac{1}{max(d(b_{i},b_{j}),1)}, if ~ i \neq j,
\end{equation}
where $W_{ij}$ (shown in bottom right of Figure \ref{methodology-table}) is transpose invariant and $W_{ij}=0$ if $i=j$.
\par
The heat kernel, $H_{t}(W)$, is a function of graph connectedness and can be used to smooth the values of each node on a graph over time, $t$.  The heat kernel is defined in the spectral domain as $g_t(\lambda) = \exp(-t \lambda)$,
    where $\lambda \in [0, 1]$ are the normalized eigenvalues of the
    graph Laplacian (formed by $W$). Since the kernel is applied to the graph eigenvalues $\lambda$, which
    can be interpreted as squared frequencies, and as a
    generalization of the Gaussian kernel on graphs. 

We apply this kernel to the object importance vector to propagate the importance of the objects based on their proximity to one another as shown
\begin{equation}
I_{P}(j)=\sum_{i=1}^{|V|} 
I_{O}(b_i) \times H_{t}(i,j),
\end{equation}
where $|V|$ is the number of connections a node has on the graph and $t$ represents the dispersion time parameter (as $t \! \! \rightarrow \! \! \infty$, all importance scores $I_P$ become uniform) set to the standard value of 1 \cite{gsp2017}. We then normalize the sum of $I_{P}$ to 1 as a post-processing step.
\par
Once we have determined the final importance weight after propagation, denoted as $I_P$, of each object, we apply a threshold $T$ in order to select a suitable subset of objects (with $I_P > T$) to be used for object evaluation. Therefore, increasing $T$ allows for more focus on the central aspects of the scene while decreasing $T$ incorporates more details from the annotations at the risk of including noisy or irrelevant annotations.
\subsection{Extending Existing Metrics}
  After our proposed approach of selecting critical objects, standard evaluation procedures are then employed. In practice, precision performance is aggregated across all object categories. Region proposals are selected based on either a confidence threshold or by taking a set number of the most confident predictions. There are 6 evaluation metrics currently employed in object detection: $mAP_{IOU=0.50:0.05:0.95}$, the primary COCO evaluation metric, $mAP_{IOU=0.50}$, the primary metric for PASCAL VOC and VL task object detection, $mAP_{IOU=0.75}$, the precision metric most attune to localization, and 3 variations of $mAR_{IOU=0.50:0.05:0.95}$ where either 1, 10, or 100 annotations are used as the ground truth set. Additional detail regarding these metrics can be found in \cite{evalsurvey2020}. For convenience, we will refer to these 6 COCO metrics as $mAP$, $mAP_{50}$, $mAP_{75}$, $mAR^{1}$, $mAR^{10}$, and $mAR^{100}$. We also include an adjusted recall metric with an IOU of 0.50 $mAR_{IOU=0.50}^{1}$ which when combined with $mAP_{50}$ via harmonic mean, creates a proposed F1 score, $F1_{IOU=0.50}$. For convenience, we will refer to our proposed metrics as $mAR_{50}^1$ and $F1$. All proposed metrics are developed based on the implementation from \cite{coco2014}. By combining the perspectives of precision and recall with the importance threshold of object annotations, our approach provides  insight into detector comparison and improvement.
\section{Experiments}
\subsection{Alignment with Captioning Metrics}
\begin{figure}[t]
\centering
\def\svgwidth{\columnwidth}
\includegraphics[width=\columnwidth,height=\textheight,keepaspectratio]{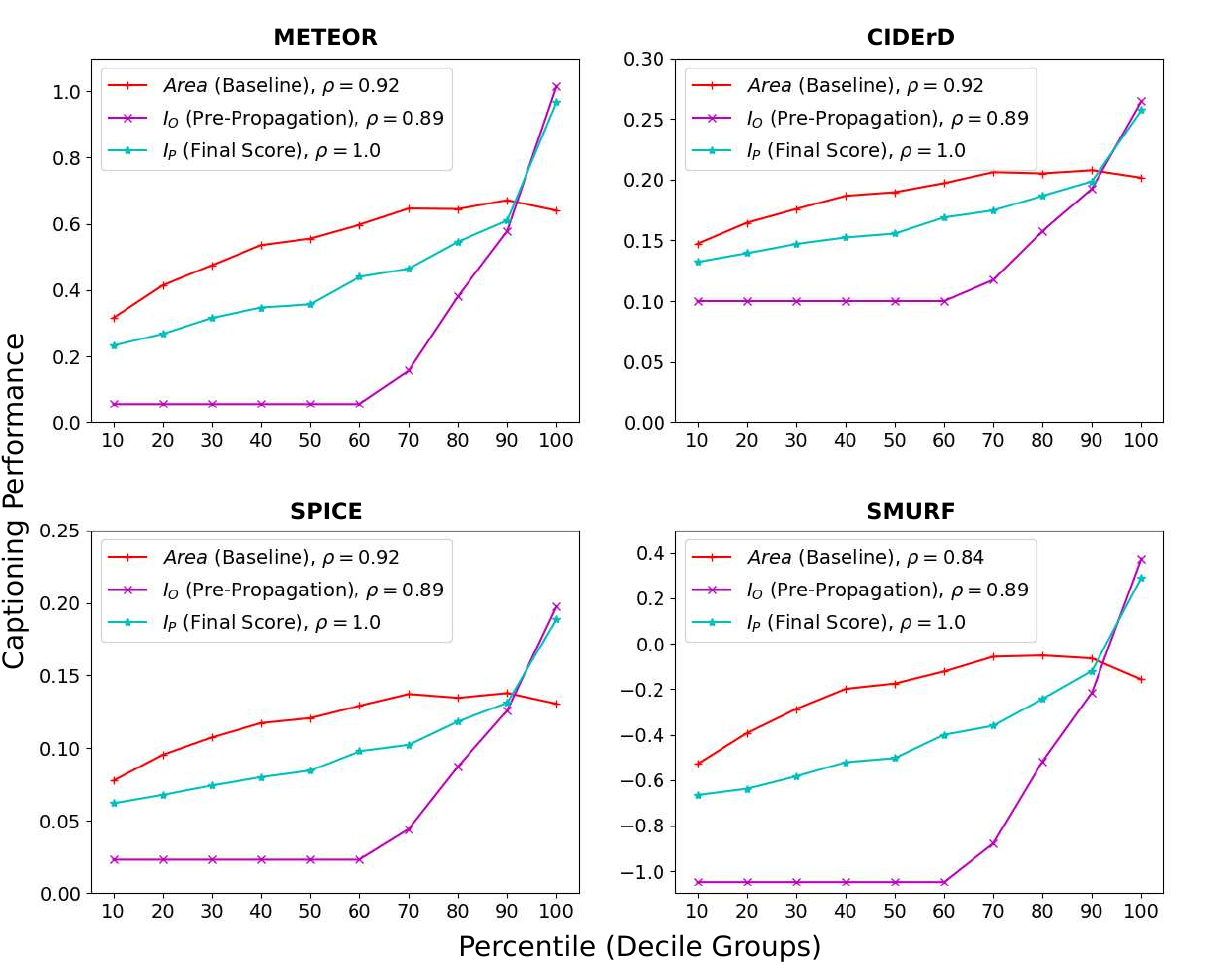}
\caption{Plots of the captioning performance for each importance score decile group for image captioning on the COCO ``Karpathy'' test split \cite{coco2014,karp2015} evaluated using CIDEr (C) \cite{cider2015}, Meteor (M) \cite{meteor2005}, SPICE (SP) \cite{spice2016}, and SMURF (SM) \cite{smurf2021}. The Spearman's $\rho$ rank correlation is used to measure the alignment between our importance score selections and image captioning results and is shown for each method in the legend.}
\label{empirical}
\end{figure}
We first measure our method's agreement with downstream captioning metrics when the importance scores are used to select annotations and proposals from VINVL \cite{vinvl2021} to perform the task of image captioning. VINVL uses the 1594 most frequent object classes and 524 most frequent object attributes from VG for their label prediction set. Their work uses OSCAR \cite{oscar2020} as a downstream captioner, which takes category tags and features from regions of interest as input and uses the CIDEr optimization methodology \cite{cideropt2017}. Annotations and proposals deemed more important by our algorithm should result in higher captioning scores, while annotations and proposals deemed less important by our algorithm should result in lower captioning scores. We select image captioning as the representative task of the VL domain since it directly incorporates text-image alignment in a consistent manner. 
\par
For our experiment, we first select the 2109 images from the 5000 images in the Karpathy test split \cite{karp2015} that have at least 1 VG annotation. We then remove 38 examples with poor image-caption alignment, leaving us with 2071 images for use as a benchmark. We use the provided pre-trained models with default settings and do not perform any fine-tuning. OSCAR is provided with regional information sourced from both ground truth annotations and VINVL \cite{vinvl2021} and tag information sourced from ground truth annotations. To measure agreement, we split the annotations into the 10 decile groups based on the $I_{P}$ importance scores to generate adjusted tag and corresponding bounding box feature sets as input to the OSCAR captioner. Figure \ref{empirical} shows the Spearman's $\rho$ rank correlation between the mean captioning metric scores and our importance score, along with the correlation for a developed area-based baseline. We select the Spearman's $\rho$ rank correlation since we expect the rank of the mean captioning score for each percentile range to increase in a monotonic, linear fashion if our importance scores are well aligned with VINVL and the human annotations. Figure \ref{main_figure} follows a similar procedure, but instead removes the lowest importance decile group at each iteration and shows the result for a data example at 3 different tag and feature removal points.
\par
Our results show that the $I_P$ importance score is highly aligned with captioning metrics. Although the max captioning scores from 90th-100th percentile of the pre-propagation and final scores are comparable, performance in the lower percentiles is very low and for the most part constant for the pre-propagation scores. This agrees with our expectations since the pre-propagation scores are sparse and provide no information about objects not mentioned in the captions. The area-based baseline is roughly correlated with improved captioning performance but has a drastically lower performance in the higher percentile groups.
\subsection{Human Surveys}
\begin{figure}[t!]
\centering
\def\svgwidth{\columnwidth}
\includegraphics[width=\columnwidth,height=\textheight,keepaspectratio]{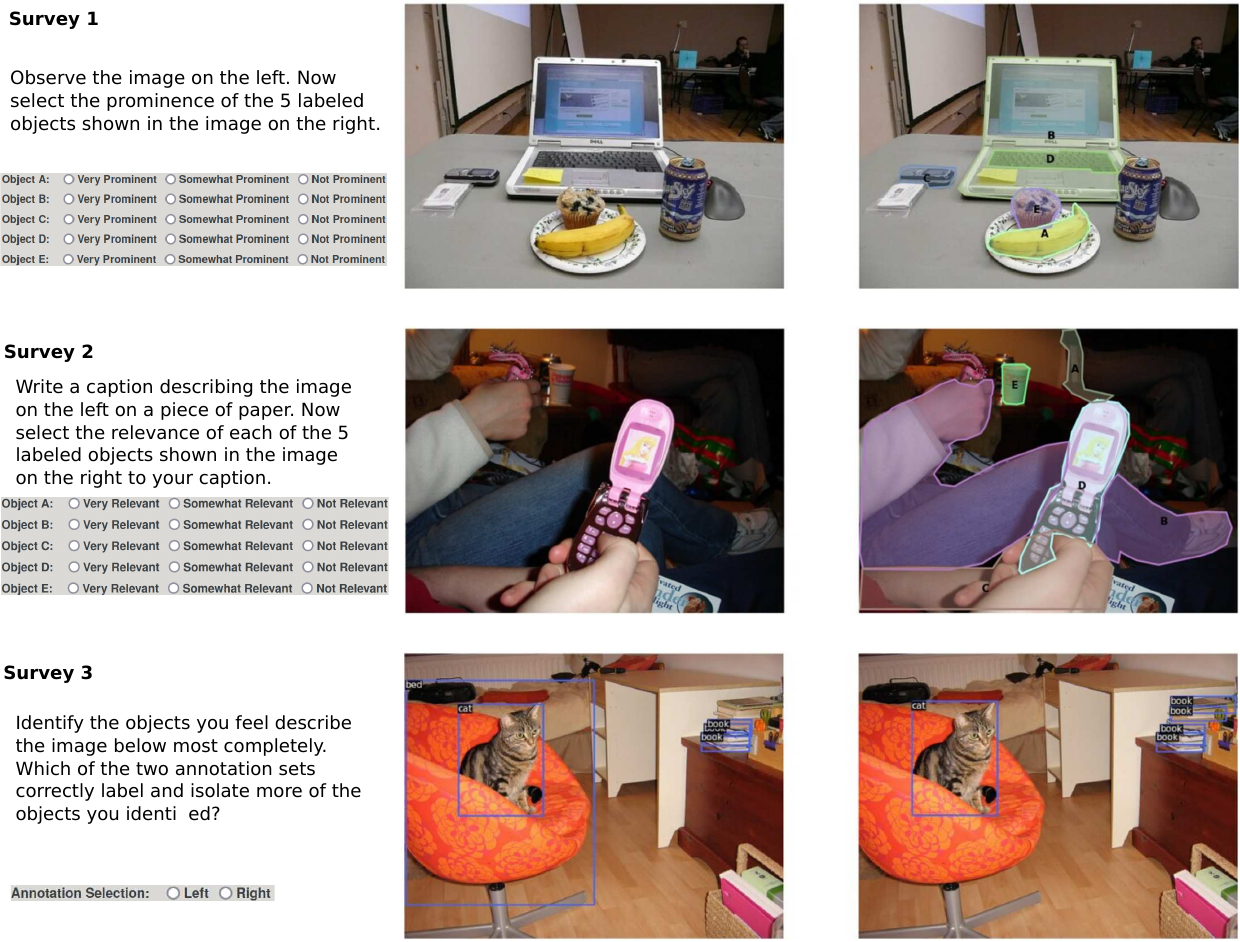}
\caption{A visualization of the 3 AMT surveys performed with the instructions and input interface shown on the left and example images shown on the right.}
\label{survey}
\end{figure}
We perform 3 human surveys using AMT in order to provide an example-level view of our approach and collect information about object importance based on human perception. We start with a set of 225 randomly selected images from COCO that contain at least 5 annotated objects. We then use the importance score of each object to select 5 objects from each image to be visually annotated using the provided ground truth regions and labelled for our first two surveys. Letter labels \{`A',`B',`C',`D',`E'\} are assigned to objects randomly and images with confusing labels due to object overlap were removed, leaving 198 labelled images that were used in our first two surveys. The first two surveys ask Turkers to rate objects based on two separate criteria in order to reduce survey bias and provide a more diverse view of object importance. For the first survey, Turkers are asked rate each object's prominence on a scale of 1 to 3, 1 being "not prominent" and 3 being "very prominent". In the second survey, we follow a similar procedure except we ask Turkers to first write a caption describing the image, then rate the objects based on their relevance to that caption. For our third survey, two state-of-the-art anchor-free detectors, FoveaBox \cite{fovea2020} and fcos \cite{fcos2019}, are used to automatically annotate our previously selected 225 images with bounding boxes and class labels. We include only the top 5 most confident predictions of each model. The 142 images with inconsistent class labels were used for the third survey, which asks Turkers to choose which image of the automatically annotated images includes and correctly labels more of the objects most important for understanding the scene. Each detector's annotated image was placed randomly on either the left or right side of the survey page for the selection process. The majority decision from 3 different Turkers is used as the final selection. The selections were quite consistent with all 3 annotators choosing the same image 81\% of the time. The full survey is shown in Figure \ref{survey}. We publicly release the AMT responses (822 in total), survey templates, and labelled images for all the surveys.
\begin{table}[t!]
\footnotesize
\centering 
\begin{tabular}{c|c|c} 
\hline
Method & Prominent & Caption Aligned\\
\hline
$Area$ (Baseline) & 0.089 & 0.064 \\
$I_{O}$ (Pre-Propagation) & 0.154 & 0.083\\
$I_{P}$ (Final Score) & 0.152 & 0.160\\
\hline
\end{tabular}
\vspace{1px}
\caption{Kendall Tau rank correlation between object scoring algorithms and rating-based survey results. 'Prominent' corresponds to the survey 1 responses while 'Caption Aligned' corresponds to the survey 2 responses. The correlation between the two surveys was 0.25. A ranking based on the area of the annotations is used as a baseline along with an ablation study.} 
\label{table:weight_correlation} 
\end{table}
\par
\begin{figure}[t!]
\centering
\def\svgwidth{\columnwidth}
\footnotesize
\begin{tabular}{c|c} 
\hline
Metric & Acc \\
\hline
$mAP$ & 0.489 \\
$mAP_{50}$ & 0.692\\
$mAP_{75}$ & 0.600\\
$mAR^{1}$ & 0.559\\
$mAR^{10}$ & 0.425\\
$mAR^{100}$ & 0.425\\
$mAR^{1}_{50}$ (Ours) & 0.750\\
$F1$ (Ours) & 0.737 \\
\hline
\end{tabular}
\\
\includegraphics[width=0.75\columnwidth,height=\textheight,keepaspectratio]{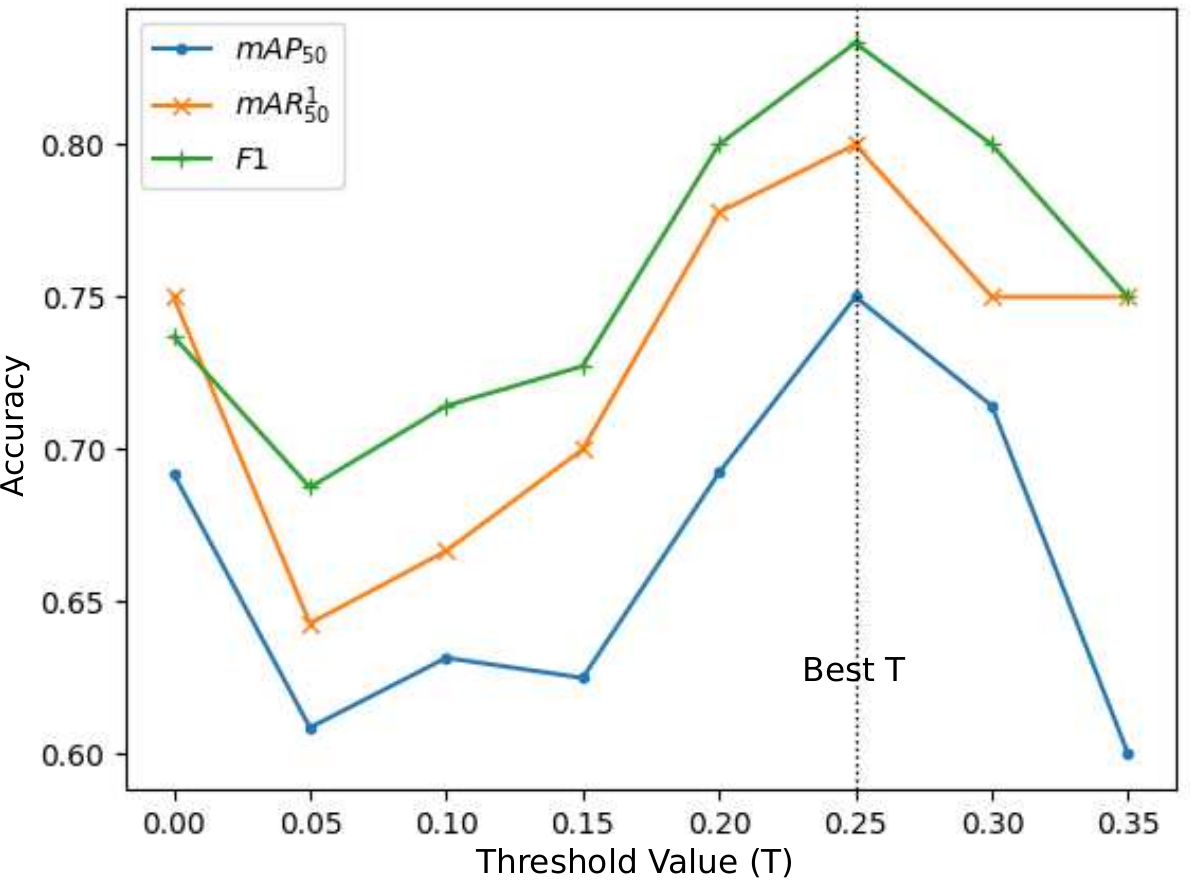}
\caption{The table on the top shows the example-level accuracy with human judgement from Survey 3 for different detection metrics using all ground truth annotations (T=0). The plot on the bottom shows how this accuracy improves for the best performing metrics by selecting ground truth object annotations based on importance with T=0.25 (92\% of annotations removed) yielding the best results.}
\label{fig:accuracy}
\end{figure}
\begin{figure}[t!]
\centering
\def\svgwidth{\columnwidth}
\includegraphics[width=0.8\columnwidth,height=\textheight,keepaspectratio]{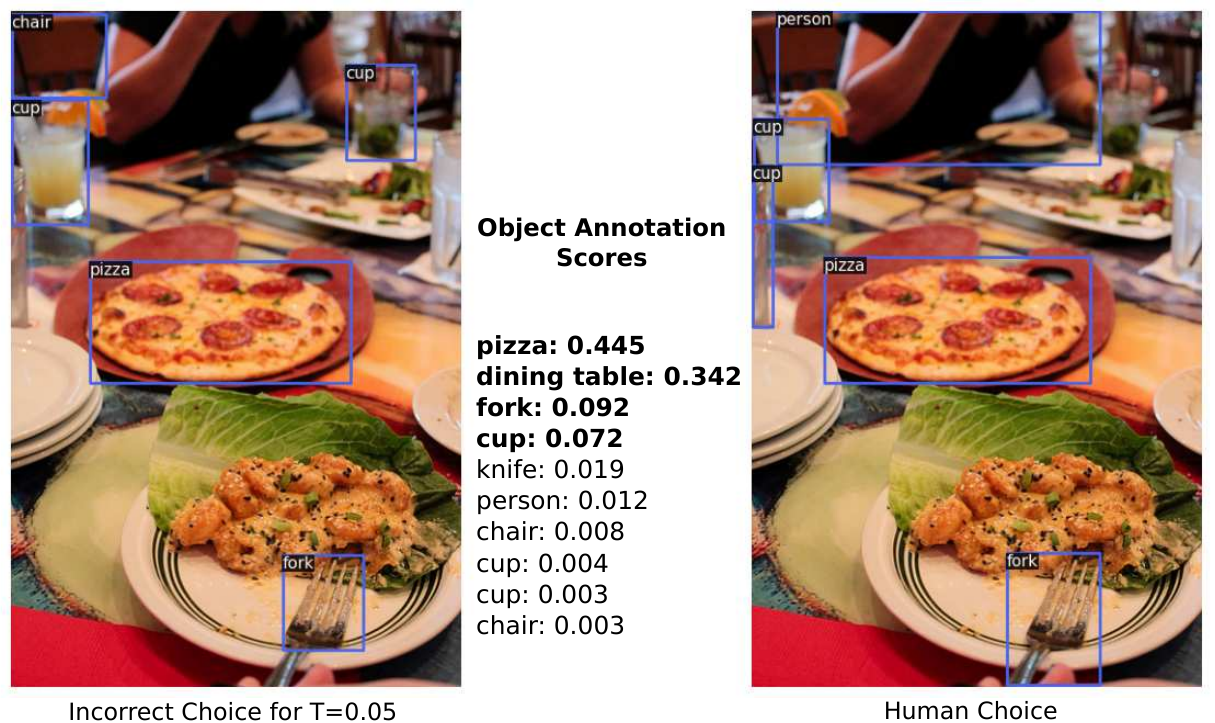}
\caption{A failure case for $mAP_{50}$ for $T\!\!=\!\!0.05$. Object annotations used in the evaluation are shown in bold.}
\label{failure}
\end{figure}

Table \ref{table:weight_correlation} shows the rank correlation of each object scoring methodology with the human responses from our first two surveys, as well as an additional ablation study showing the importance of propagating the importance over regions yielding results consistent with Figure \ref{empirical}. As a baseline, we use object area as a method for importance scoring. For the purposes of this comparison, the importance scores are mapped to the discrete set \{1,2,3\} such that the frequency of each value matches that of the human survey response distribution. We use Kendall Tau due to its focus on concordant and discordant pairs, making it more robust to ties and survey noise and more appropriate for experiments not fitting to a linear representation. Our importance scores demonstrate a dramatic increase in human object rating alignment over the more na\"ive area-based approach for both survey prompts, despite the inter-correlation between the surveys being only 25\%. Based on the results from the third survey, we measure the agreement between our proposed detection evaluation metrics along with existing metrics in the table in Figure \ref{fig:accuracy}. We observe that $mAP_{50}$ along with our proposed recall and F1 scores have the greatest alignment with human judgement when compared with other existing metrics. We are able to further improve alignment with human judgement by using our annotation selection methodology, which can be seen in the plot shown in Figure \ref{fig:accuracy}. The initial dip in alignment between human and metric selections is largely caused by the forced selection of object rankings in Survey 1 and 2 rather than allowing for ties. This necessary limitation on the human survey's granularity simplifies the response process but forces arbitrary selections for lower importance objects. One such example can be observed in Figure \ref{failure}. This selection should have instead been considered a tie by the metric since most scene-essential objects have been accounted for by both models and any image selection is likely to be arbitrary.
\par
\subsection{Consistency Study}
Although COCO is a crucial benchmark for object detection, VG is one of the primary benchmarks for VL tasks. Therefore, we assess whether our approach is consistent across the COCO and VG annotation formats. By showing the consistency of selected objects across datasets, we further support the existence of a critical subset of ground truth object annotations and the effectiveness of our approach.
\par
\begin{figure}[t!]
\centering
\def\svgwidth{\columnwidth}
\includegraphics[width=0.75\columnwidth,height=\textheight,keepaspectratio]{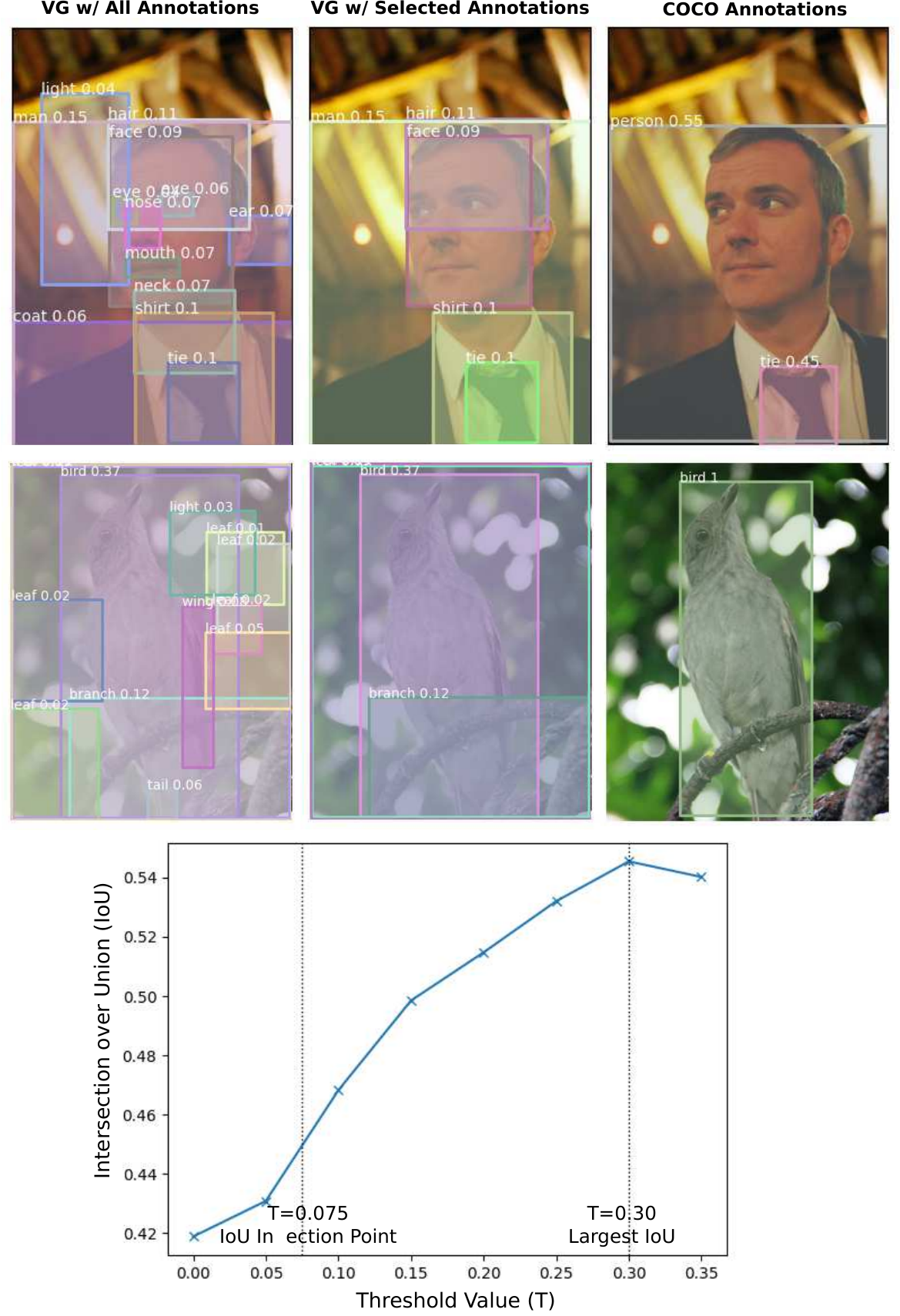}
\caption{Examples of the annotation selection process for Visual Genome using a threshold value of 0.075 are shown on the top along with a comparison to COCO. The plot on the bottom shows the average IOU between COCO and VG annotations as the importance threshold is raised.}
\label{vg-generalize}
\end{figure}

In place of captions, VG utilizes region descriptions where a human annotator describes a section of the image. These descriptions are very similar to captions and at least 10 descriptions are provided for each image, so we postulate they can be used in place of captions for our method. We use the preprocessed VG split from \cite{sgbase2017}, commonly utilized for SGG tasks, which contains information on only the 150 most frequent object categories, to perform a study. We first acquire the annotations of 11,597 training split images that appear in both the VG and COCO dataset. Then we perform our score-based selection of objects on the COCO dataset with a threshold value of 0.25. We perform the same selection on VG, but sweep the threshold value from 0 to 0.35 in increments of 0.05. Here we use the Intersection-over-Union (IOU) between the selected annotations of the two datasets to show that our method is gradually removing excessive and noisy annotations and instead focusing on essential regions, like those annotated in COCO, as supported by the results in Figure \ref{vg-generalize}. Visualizations of selected annotations for two images show that selected VG annotations become very similar to the COCO annotations in terms of quality and focus, but still include some additional detail. The gradual increase in IOU is highly significant since it occurs despite the large amount of overlap between the annotations of VG. Our results suggest that some of the additional annotations found in VG actually lead to worse scene coverage and that there is a subset of objects recognized by annotators as capturing the essence of the scene, which we are able to identify with our method. Based this study, we determine T=0.075 (61\% of annotations removed) to be a threshold of interest for the VG dataset since an inflection point occurs around this value, meaning many of the most irrelevant annotations have been removed and the rate of IOU increase has begun to slow. We also determine T=0.30 (96\% of annotations removed) to be a threshold of interest since this selection has the greatest IOU with the selected COCO annotations.
\subsection{Importance in a SGG Benchmark}
Current VL works typically report the $mAP_{50}$ score of a proposed detector as a intermediate validation of their methodologies. While $mAP_{50}$ is certainly correlated with the bias needed for improved VL performance, cases of disagreement between the intermediate measure and downstream task performance measures are very common. Another issue is that since specialized detectors like VINVL \cite{vinvl2021} and BUTD \cite{topbottom2018} use unique category sets and training procedures, it is difficult to compare them with other detectors. Scene Graph Generation (SGG) benchmarks, on the other hand, have a strict category set and training procedure used by proposed detectors, allowing for direct and meaningful comparison. Authors have also claimed that SGG can serve as an important benchmark for connecting detection and scene-understanding tasks. However, there is little to no evidence to support that SGG provides any more feedback on VL-oriented performance than $mAP_{50}$, nor is there any evidence that techniques improving SGG performance lead to improved performance in other VL tasks. On the contrary, the novel portion of SGG, the object relations, have been found to be highly imbalanced and ambiguous \cite{unbiased2020}. In addition, it is difficult to determine how results reported in Scene Graph Generation works can be related to other VL tasks. %
\par
In this experiment, we present a use case that highlights the flaws of these competing methodologies in a highly explainable manner. We follow the progress of detectors utilized in Scene Graph Generation (SGG) by extensively evaluating 3 selected detectors. All detection models in SGG currently use the Faster RCNN architecture \cite{frcnn2015}. However, the 2018 model from ``Neural Motifs'' \cite{motifs2018} incorporates a VGG backbone \cite{vgg2015} and both the 2020 model from \cite{unbiased2020} and the 2021 model \cite{sgg2021} incorporate the ResNeXt101-FPN as the backbone as originally done in \cite{graphical2019}. Since VG has approximately 20 object annotations per image, we use the top 20 most confident proposals from each detector in our evaluation. A comparison of the performance of these detectors is shown in Table \ref{table:detector_compare}.
\par
Based on this analysis, it is apparent that the information provided by previous measures is very limited. Questions like which objects were detected and whether these objects were worth detecting are ignored. 
Using our more detailed evaluation, we see the 2021 detector consistently achieves higher recall than the other two detectors, but the 2018 model has higher precision on objects critical to VL tasks. This leads to the 2018 model having a larger F1 score at the higher importance threshold and shows that newer models may not be adequately prioritizing the correct classification of essential objects, a phenomenon not captured by existing evaluation methods due to their misalignment. While our measures demonstrate a degree of agreement with previous methods, our method is able to capture a much broader and more nuanced story about a given detector's performance.
\begin{table}[t!]
\footnotesize
\centering 
\begin{tabular}{c|c|c|c|c|c|c|c}
\hline
& Previous &\multicolumn{6}{c}{Ours} \\
\hline
\multirow{2}{*}{\!\!Model} & $T\!\!=\!0$ & \multicolumn{3}{c|}{$T\!\!=\!0.075$} & \multicolumn{3}{c}{$T\!\!=\!0.30$}\\
& $P$ & $P$ & $R$ & $F$ & $P$ & $R$ & $F$\\
\hline
2018 \cite{motifs2018} & 20.4 & 18.0 & 37.7 & 24.3 & 5.9 & 46.7 & 10.6\\
2020 \cite{unbiased2020} & 22.9 & 18.7 & 40.0 & 25.5 & 5.2 & 47.2 & 9.4 \\
2021 \cite{sgg2021} & 24.5 & 20.0 & 41.7 & 27.0 & 5.7 & 50.9 & 10.2\\
\hline
\end{tabular}
\vspace{1px}
\caption{Detector evaluation based on the Visual Genome experimental procedure from \cite{sgbase2017}. The ``Previous'' column represents the relevant information provided by the unmodified $mAP_{50}$ metric, while the remaining columns correspond to the feedback provided by our proposed modifications using thresholds determined in the consistency study.} 
\label{table:detector_compare} 
\end{table}
\section{Discussion and Limitations} 
It should be noted that selection of critical objects is by no means a ``one size fits all solution'' to metric misalignment. Our method is intended to be an enhancement of existing VL evaluation metrics like caption evaluation by providing more detailed feedback on object prioritization by the vision module based on downstream semantic information. Furthermore, proper alignment between object proposals and evaluation is a task-specific problem where the severity of misalignment will vary depending on how few of the annotated objects are relevant to a given downstream task. 
\par
Another challenge facing VL evaluation metrics is degrading performance when there are fewer or noisier captions provided by human annotation \cite{spice2016}. For our method, fewer captions would result in more data examples being skipped during the evaluation due to no importance being assigned to any of the objects in the image. We also find that while the mappings performed by ConceptNet are very reasonable, there are a number of failure cases. An example of such a case is the word ``player'', which is mapped to ``sports ball'' instead of the more appropriate ``person''. This can lead to sports players being given less importance at the $I_{O}$ stage, especially if all the annotators only use the term ``player'' to describe the person in question. Importance propagation helps compensate for this issue as long as the ``player'' is in close proximity to the ``sports ball''.
\par
We also observe some limitations in the human study in the form of annotation noise. We attempt to reduce influence of this noise by using two different prompts for the first two surveys, using the majority decision of 3 responses in the third survey, and removing select responses. To avoid bias in our selections, we do not keep any intermediate data from the survey process and instead make removal selections based on survey completion in an unreasonable amount of time, with too little variation in scoring, or with excessively repeated patterns. In addition, there is a clear trade-off between instruction specificity and bias when conducting surveys \cite{blame2022}. We appeal to the AMT annotators innate understanding of importance with unique and open-ended tasks, but this can potentially lead to less consistent responses. 
\section{Conclusion and Broader Impact}
We present a formulation of misalignment in object proposal metrics along with an importance score that can be used to select objects critical to VL tasks in order to address this phenomenon. Our object proposal evaluation methodology is the first to be validated with both human judgement and empirical results. The current lack of a VL specific detector evaluation benchmark has contributed to a shift towards embedding-based approaches for VL tasks \cite{embedding2020,embedding2022}. This shift represents a dangerous trend in VL pipelines, which are known to be sensitive to language \cite{capbias2018} and dataset \cite{vqamatter2017} priors. To avoid such issues, vision and language components of these pipelines should be evaluated independently in an explainable manner. Our approach could help revitalize research in detectors and enable transparent and explainable approaches and analyses in VL tasks. Future work could focus on applying our method to other elements of scene understanding such as activities or attributes by shifting the focus of the concept extraction to the element of interest.

\noindent {\bf Acknowledgments: }
The authors acknowledge support from the NSF RI Project VR-K \#1750082, CPS Project \#2038666 and and a grant from Meta AI Learning Alliance. Any opinions, findings, and conclusions in this publication are those of the authors and do not necessarily reflect the view of the funding agencies.

{\small
\bibliographystyle{ieee_fullname}
\bibliography{egbib}
}
\end{document}